\documentclass[runningheads]{llncs}

\usepackage{graphicx}
\usepackage{mathtools}
\usepackage{subfig}
\usepackage{changepage}
\usepackage[dvipsnames]{xcolor}
\usepackage{enumitem}
\usepackage{footnote}
\makesavenoteenv{tabular}
\makesavenoteenv{table}

\title{Enhancing Fault Tolerance of Neural Networks for Security-Critical Applications}

\titlerunning{Enhancing Fault Tolerance of Neural Networks}

\author{Manaar Alam\inst{1} \and Arnab Bag\inst{1} \and Debapriya Basu Roy\inst{1} \and Dirmanto Jap\inst{2} \and Jakub Breier\inst{3} \and Shivam Bhasin\inst{2} \and Debdeep Mukhopadhyay\inst{1}}

\authorrunning{M. Alam \emph{et al.}}

\institute{
Indian Institute of Technology Kharagpur, India\\
\email{\{alam.manaar, amiarnabbolchi, dbroy24, debdeep.mukhopadhyay\}@gmail.com}
\and
Nanyang Technological University, Singapore\\
\email{\{djap, sbhasin\}@ntu.edu.sg}
\and
Underwriters Laboratories, Singapore\\
\email{jbreier@jbreier.com}
}

\let\oldmaketitle\maketitle
\renewcommand{\maketitle}{\oldmaketitle\setcounter{footnote}{0}}

\begin{document}

\maketitle

\begin{abstract}
Neural Networks (NN) have recently emerged as backbone of several sensitive applications like automobile, medical image, security, etc. NNs inherently offer Partial Fault Tolerance (PFT) in their architecture; however, the biased PFT of NNs can lead to severe consequences in applications like cryptography and security critical scenarios. In this paper, we propose a revised implementation which enhances the PFT property of NN significantly with detailed mathematical analysis. We evaluated the performance of revised NN considering both software and FPGA implementation for a cryptographic primitive like AES SBox. The results show that the PFT of NNs can be significantly increased with the proposed methodology.

\end{abstract}

\section{Introduction}
We have seen an outburst of research on Neural Networks (NNs) in both industry and academia over the years because of its compelling performance in wide varieties of domains, starting from image recognition, speech processing to several sensitive applications like medical diagnosis and security. In addition to the competent performance, one of the most intrinsic features of NNs is that it exhibits ``some degree" of robustness with the ability to function correctly even after the faults in any of its parameters in the architecture, which promotes its popularity in diverse applications. Fault tolerance of NNs has been extensively studied in the past by many researchers~\cite{segee1991fault,minnix1992fault,neti1992maximally,protzel1993performance,murray1993synaptic,murray1994enhanced,phatak1995complete,tchernev2005investigating,tchernev2005perfect,dos2018analyzing}. It has been demonstrated that a typical NN architecture where each neuron computes the weighted sum of inputs from other neurons cannot achieve complete fault tolerance~\cite{phatak1995complete,tchernev2005perfect}, rather achieves a \emph{Partial Fault Tolerance (PFT)}. The fault tolerance of the NNs can be increased significantly by injecting noise during the training process~\cite{murray1993synaptic,murray1994enhanced}, replication and majority voting during the prediction operation~\cite{protzel1993performance,phatak1995complete}, or by increasing the size of the network~\cite{segee1991fault,neti1992maximally,tchernev2005investigating}. Fault tolerance of cryptographic algorithms has emerged as an integral property with the advancement of fault attacks. As an example, Advanced Encryption Standard (AES), which is secure against known theoretical cryptanalysis, can be broken by a single fault injection~\cite{ali2011improved}.

\begin{figure}[!t]
    \centering
    \includegraphics[width=0.5\linewidth]{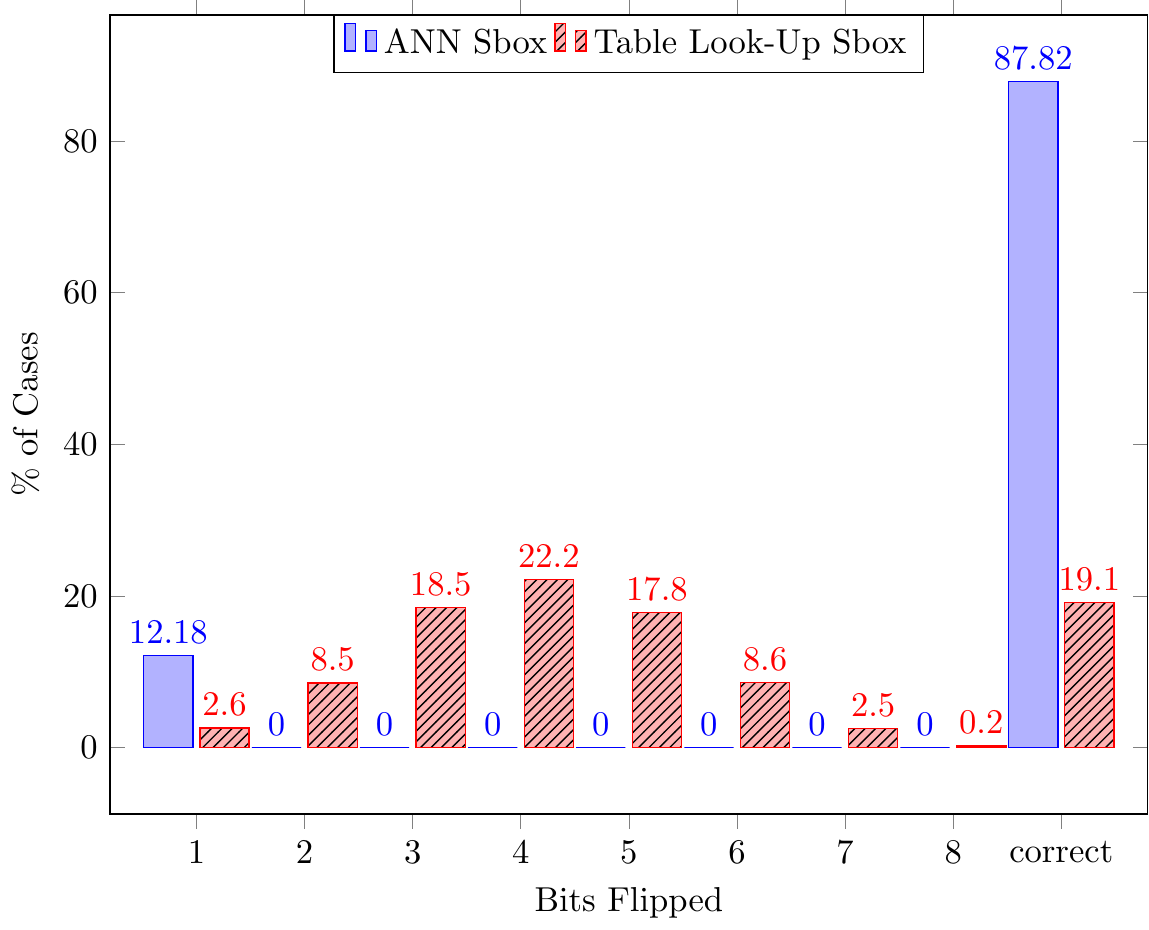}
    \caption{Comparing fault tolerance of AES SBox when implemented with Table-Lookup against NN based SBox}
    \label{fig:aes_avr}
\end{figure}

While general applications are prone to unintentional faults, cryptographic applications suffer from targeted faults from a powerful adversary. These faults can be injected intentionally by creating a hazardous environment (like laser injection, clock and voltage glitch, high-intensity electromagnetic wave injection, etc.) around the system. A fault-tolerant architecture for a cryptographic application will thwart such kind of attacks. In our case study, we have concentrated on AES as it is the de facto standard of block ciphers around the world. One of the primary components in AES encryption, where the secret key is directly involved in the computation, is the \texttt{AddRoundKey} operation. The non-linearity in the calculation is induced by the \texttt{SubBytes} operation, using SBox mapping. To give a snapshot of the problem at hand, we performed a practical fault injection on $f=SBox(x\oplus k)$ where $x$ is a byte input and $k$ is one byte of secret key, running on a commercial off-the-shelf microcontroller. Fig.~\ref{fig:aes_avr} compares fault tolerance of $f$ implemented using look-up table and boolean logic against the same function implemented with a NN (having 8 neurons in the input layer, 2 hidden layers with 400 and 200 neurons, and one output layer with 8 neurons) for the mapping of 8-bit input to 8-bit output. Faults were achieved by laser fault injection technique, aiming at control-flow disturbance. Injection time varied randomly, covering the entire computation of $f$, while the other laser parameters were fixed to guarantee the faults. The NN parameters were chosen to have 100\% accuracy with no constraints on fault tolerance. It can be observed that a blindly chosen network resulted in a fault tolerance of 87\%.

Motivated by the above result, in this paper, we propose to exploit the PFT property of NN for cryptographic applications. As a proof of concept, we implement combined AES SBox preceded by \texttt{XOR} with secret key over one byte ($f=SBox(x\oplus k)$), hereafter referred as $f$ operation, using NNs which exhibits significantly higher degree of fault tolerance compared to the standard designs. We also propose a method to further improve the PFT of the implementation by incorporating several constraints on the NN parameters during the training process. Later, implementation overhead of the developed highly fault tolerant NN architecture for $f$ is also reported. The implementation results are reported on Xilinx Artix-7 FPGA (Basys 3 board), which allows us enough flexibility to test different settings varying from low overhead, less fault-tolerant to high-overhead, highly fault tolerant designs, against typical microcontroller.

\subsection*{Contribution}
The primary \textbf{contributions} of this work are:
\begin{enumerate}
\item We present an analytical way to develop a highly fault tolerant cryptographic primitive like AES SBox using an NN having a much higher degree of fault tolerance than the standard implementation. To the best of our knowledge, this is the first work to achieve a highly fault-tolerant AES SBox which has only $1.21 \times 10^{-5}\%$ chance of producing faulty outputs even after inducing faults in network parameters.
\item We have implemented the fault tolerant NN architecture in an FPGA platform with tailored implementation strategies. The modified implementation, as it enforces the fault-tolerance property, incurs extra implementation overhead for both software and hardware.
\end{enumerate}

\section{Preliminaries on Neural Network}\label{sec:prelim}
In this section, we discuss elementary Neural Network (NN) terminologies which has been used throughout this paper. We have considered a basic NN architecture with three layers - the \emph{Input} layer, the \emph{Hidden} layer, and the \emph{Output} layer, which contain $l$, $m$, and $n$ neurons respectively. The \emph{Activation} functions for the hidden layer and the output layer are \emph{ReLU} and \emph{Softmax}, respectively. The network is trained using standard \emph{Gradient Descent Backpropagation} algorithm. The definition of each symbol related to the NN used in this paper are mentioned in Table~\ref{tab:definition}.

\begin{table}[!t]
\centering
\caption{Definition of symbols used throughout this paper}
\label{tab:definition}
\begin{tabular}{|c|l|}
\hline
\multicolumn{1}{|c|}{\textbf{Symbols}} & \multicolumn{1}{c|}{\textbf{Definition}} \\ \hline
$x_i$ & Input to the $i^{th}$ neuron in the \emph{Input} layer \\ \hline
$h_j$ & Output of the $j^{th}$ neuron in the \emph{Hidden} layer\\ \hline
$y_k$ & Output of the $k^{th}$ neuron in the \emph{Output} layer\\ \hline
$w_{ij}^{(1)}$ & \begin{tabular}[c]{@{}l@{}}Weight of the \emph{link} connecting $i^{th}$ neuron in the \emph{Input} layer\\ with the $j^{th}$ neuron in the \emph{Hidden} layer\end{tabular}\\ \hline
$w_{jk}^{(2)}$ & \begin{tabular}[c]{@{}l@{}}Weight of the \emph{link} connecting $j^{th}$ neuron in the \emph{Hidden} layer\\ with the $k^{th}$ neuron in the \emph{Output} layer\end{tabular}\\ \hline
$b_{j}^{(1)}$ & Bias of the $j^{th}$ neuron in the \emph{Hidden} layer\\ \hline
$b_{k}^{(2)}$ & Bias of the $k^{th}$ neuron in the \emph{Output} layer\\ \hline
$ReLU$ & Activation function, given by,
    $ReLU(h_j)=$
    \(\displaystyle
    \begin{cases}
      0, & \text{if}\ h_j \le 0 \\
      h_j, & \text{otherwise}
    \end{cases} \)\\ \hline
$Softmax$ & Activation function, given by,
$Softmax(y_k) = \frac{e^{y_k}}{\sum_{k=1}^{n}e^{y_k}}$\\ \hline
$Linear$ & Activation function, given by,
    $Linear(y_k)=y_k$
\\ \hline
\end{tabular}
\end{table}

The value of the $k^{th}$ neuron at the output layer is given by:
\begin{equation}\label{eq:softmax}
    y_k = Softmax(\sum_{j=1}^{m}h_jw_{jk}^{(2)}+b_k^{(2)})
\end{equation}

The value of $j^{th}$ neuron at the hidden layer is given by:

\begin{equation}\label{eq:relu}
    h_j = ReLU(\sum_{i=1}^{l}x_iw_{ij}^{(1)}+b_j^{(1)})
\end{equation}

\section{Fault Tolerant AES S-Box Design}
In this section, we use the NN architecture as discussed in Section~\ref{sec:prelim} to model the $f$ operation. We have considered only those networks which produce \textbf{100\%} classification accuracy with the training dataset, since a wrong classification results into incorrect ciphertext. The main focus in the following subsections is to investigate the PFT of the modelled NN by also concentrating on the hardware implementation of such design.

\subsection{Dataset and Network Topology}\label{sec:dataset_topology}
The NN in our experiment learns the $f$ operation. Hence, the input to the network $NN_k$ for a fixed secret key-byte $k$ is a plaintext byte $x$ and the output $y = SBox(x \oplus k)$. We have used binary bit patterns for the input $x$ and \emph{one-hot encoding} for the output $y$ in order to enhance the prediction accuracy of the model. Hence, the input layer contains $8$ neurons and the output layer contains $256$ neurons. We have experimented with different number of hidden layer neurons and the effect of number of neurons on PFT is discussed in Section~\ref{sec:design_standard}. The NN model trained using the gradient descent algorithm produces floating point values for weights and bias parameters. In order to alleviate the implementation effort on hardware, we considered integral values of all the learned parameters after the training by considering different bit precision after the decimal. The effect of selecting bit precision on PFT is also discussed in Section~\ref{sec:design_standard}. \emph{Softmax} activation function in the output layer needs computation of exponentiation as shown in Table~\ref{tab:definition}. In order to reduce the implementation complexity, we have considered \emph{Linear} activation function (also shown in Table~\ref{tab:definition}) in the output layer instead of \emph{Softmax} function while implementing the learned model in the hardware\footnote{The motivation of using linear activation function is that the output of softmax function is directly proportional to the output of linear function which can be implemented with integer parameters with less computational resources, unlike softmax function.}. Hence, during the classification process in the hardware implementation, Equation~\eqref{eq:softmax} is converted to

\begin{equation}\label{eq:linear}
    y_k = \sum_{j=1}^{m}h_jw_{jk}^{(2)}+b_k^{(2)}
\end{equation}

Decision on the correct class is taken by the \texttt{ArgMax} function on output layer neurons to find the neuron having the maximum value.

\subsection{Fault Model}\label{sec:fault_model}
Appropriate selection of a fault model for an adversary is crucial in order to evaluate the fault tolerance capability of the NN and practicality of the attack. In all our experiments the fault models that we consider are:
\begin{enumerate}
\item We assume that the learning phase is fault-free. Faults can only be injected during the classification phase, not during training phase.
\item We consider \emph{single location} fault model\footnote{Fault at a single location is actually a very realistic fault model as most of the differential fault attack works when only a byte or a nibble gets corrupted. Injecting faults at multiple location often generates faulty output which can not be exploited.}, i.e., an adversary can induce fault at only one weight or bias parameter in an individual execution.
\item An adversary can employ any of the possible fault injection methods like single-bit flip, multiple-bit flips or zero/random values.
\end{enumerate}

We have the analyzed the fault tolerance with all the possible faulty values for all the weight and bias parameters in our experiments.
In the following subsection we present the degree of PFT of NN topology with the dataset mentioned in Section~\ref{sec:dataset_topology}. We also present a rational behind the selection of bit precision after the decimal points of the floating point weights and bias values in order to convert into integer weights along with number of neurons in the hidden layer with experimental results.

\subsection{Fault Tolerance of AES SBox learned with Standard Neural Network}\label{sec:design_standard}
We have used \texttt{Keras} library to implement the NN as mentioned in Section~\ref{sec:dataset_topology}. We have trained the model to achieve \textbf{100\%} classification accuracy, and used different values of precision to represent integer parameters and number of hidden layer neurons in order to achieve a final model with satisfactory PFT. The \% Faults in our experiments is calculated as:

\begin{adjustwidth}{-1cm}{-1cm}
\centering
\begin{equation*}
    \% Faults = \frac{\#\text{Faulty Output}}{\#\text{Parameters} \times \#\text{All Possible Faulty Values of Each Parameter} \times \#\text{Inputs}}
\end{equation*}
\end{adjustwidth}

\begin{figure}[!t]
	\centering
	\subfloat[First Layer Weight]{
		\includegraphics[width=0.5\textwidth]{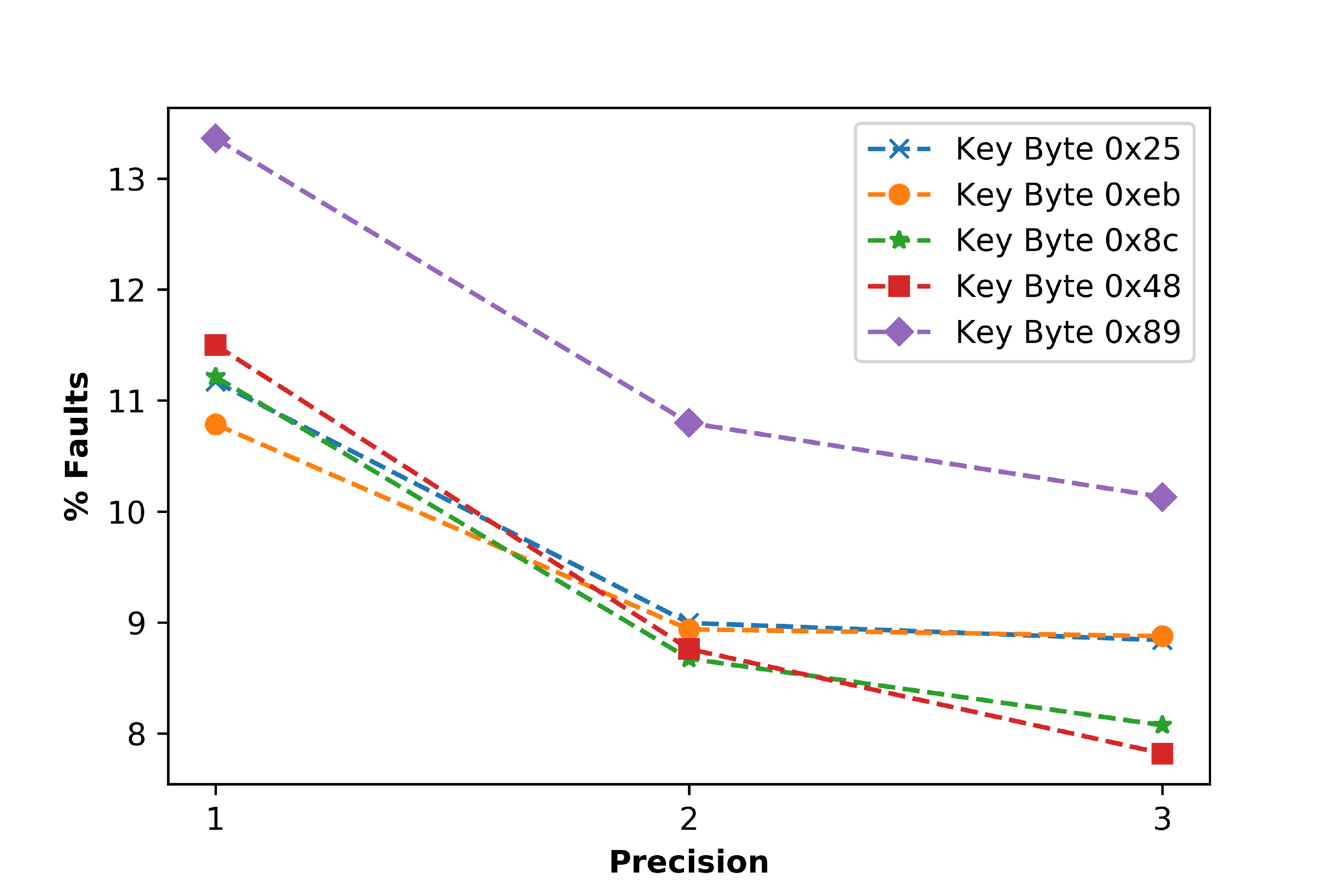}
	}
	\subfloat[Second Layer Weight]{
		\includegraphics[width=0.5\textwidth]{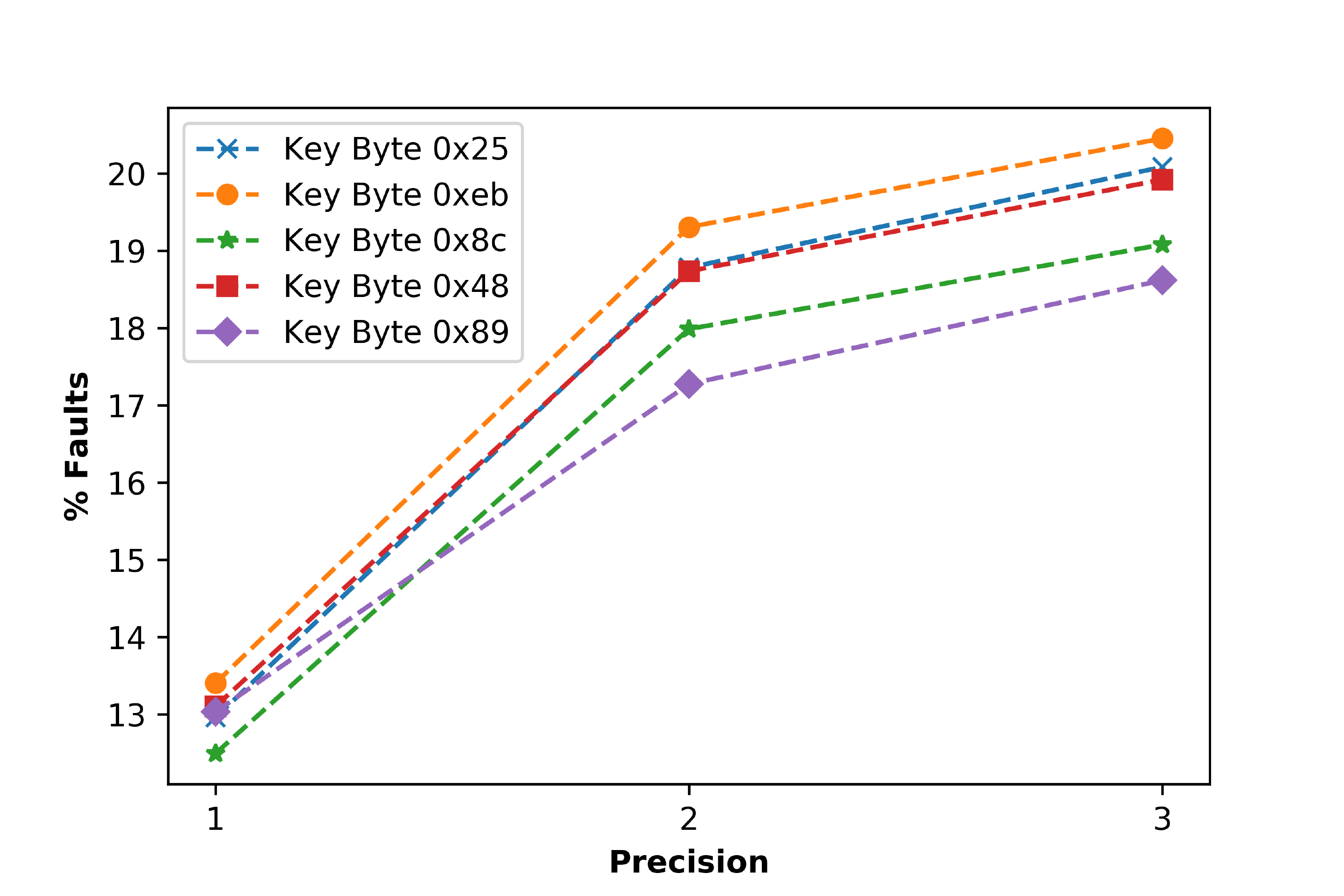}
	}
	\qquad
	\subfloat[Hidden Layer Bias]{
		\includegraphics[width=0.5\textwidth]{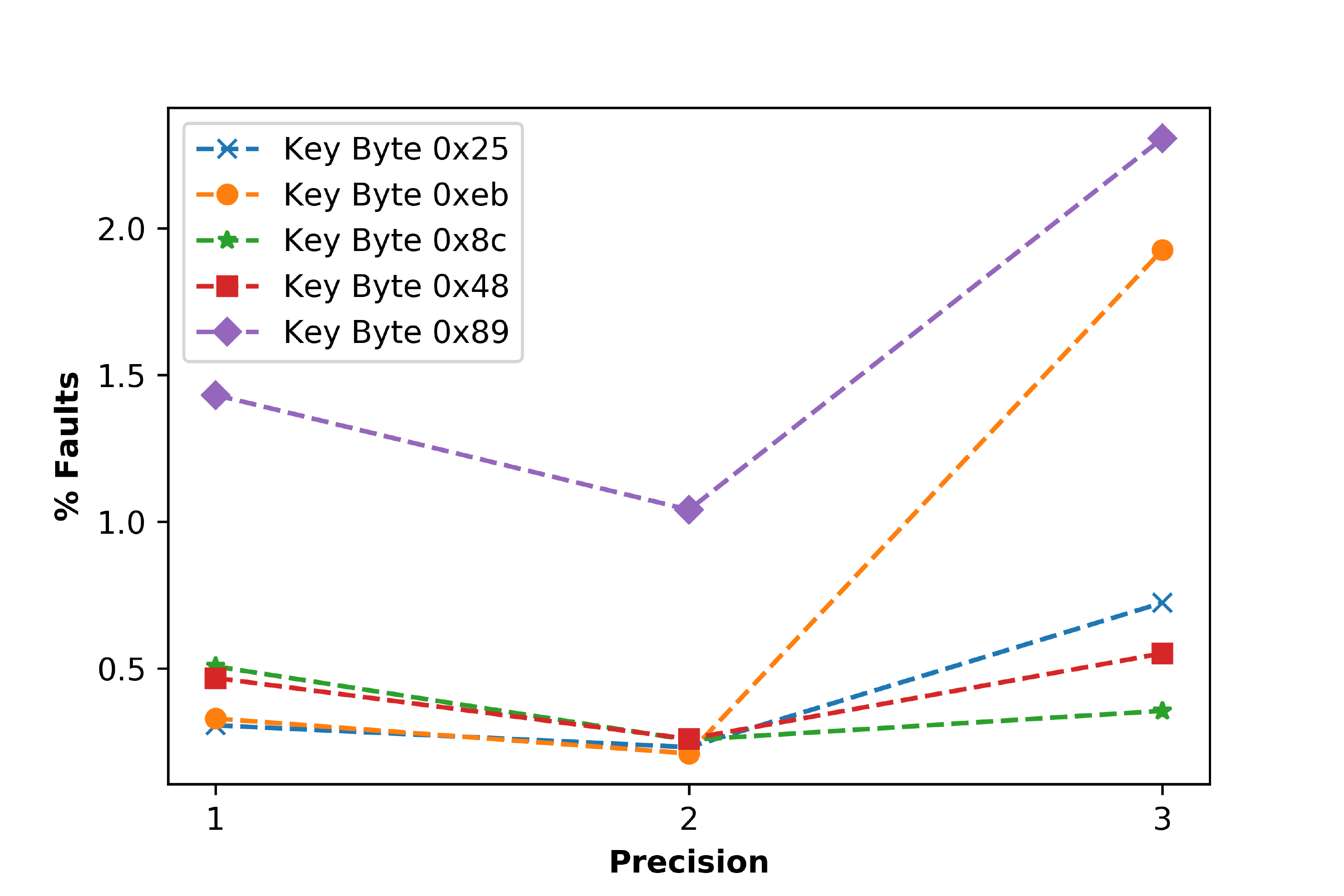}
	}
	\caption{Effect of Selecting Precision on Fault Tolerance\label{fig:fault_precision}}
\end{figure}

Fig.~\ref{fig:fault_precision} presents \% Faults for five different NNs trained with five different secret key bytes when faults are injected in $\forall_{i, j} w_{ij}^{(1)}$ (First Layer Weights), $\forall_{j, k} w_{jk}^{(2)}$ (Second Layer Weights), and $\forall_{j} b_{j}^{(1)}$ (Hidden Layer Bias)\footnote{We do not provide plots for $b_{k}^{(2)}$ (Output Layer Bias) as we have found out that these parameters are totally fault tolerant.} considering integer parameters with different precisions after the decimal point of floating point values. We can easily observe that with the increase in precision \% Faults for First Layer Weights gets decreased while for Second Layer Weights and Hidden Layer Bias it increases. We have considered 1-bit precision after the decimal points of the floating point values while converting to integer for the further implementation as it not only provides the least overall \% Faults of the NN but also helps in the implementation.

\begin{figure}[!t]
	\centering
	\subfloat[First Layer Weight]{
		\includegraphics[width=0.5\linewidth]{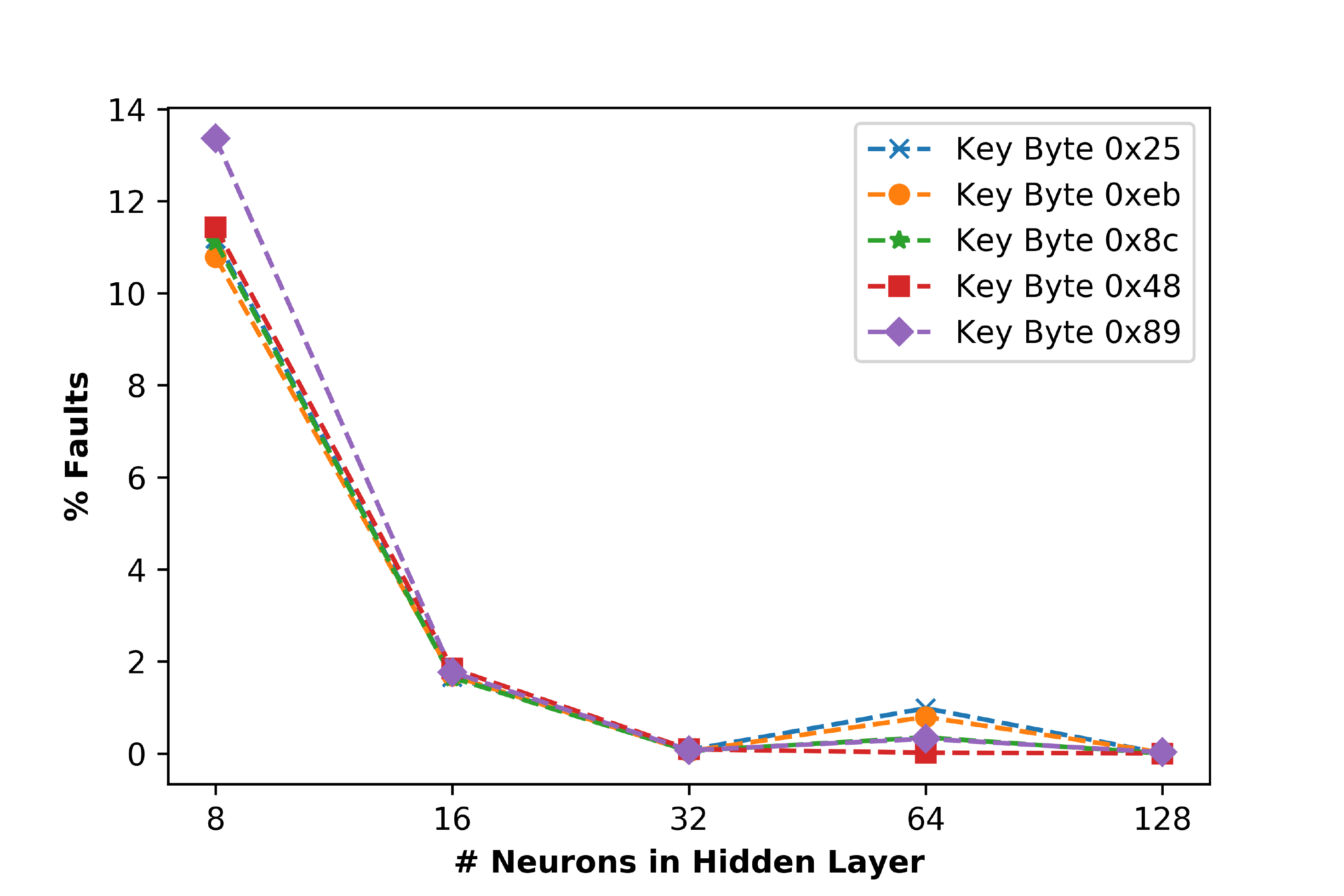}
	}
	\subfloat[Second Layer Weight]{
		\includegraphics[width=0.5\linewidth]{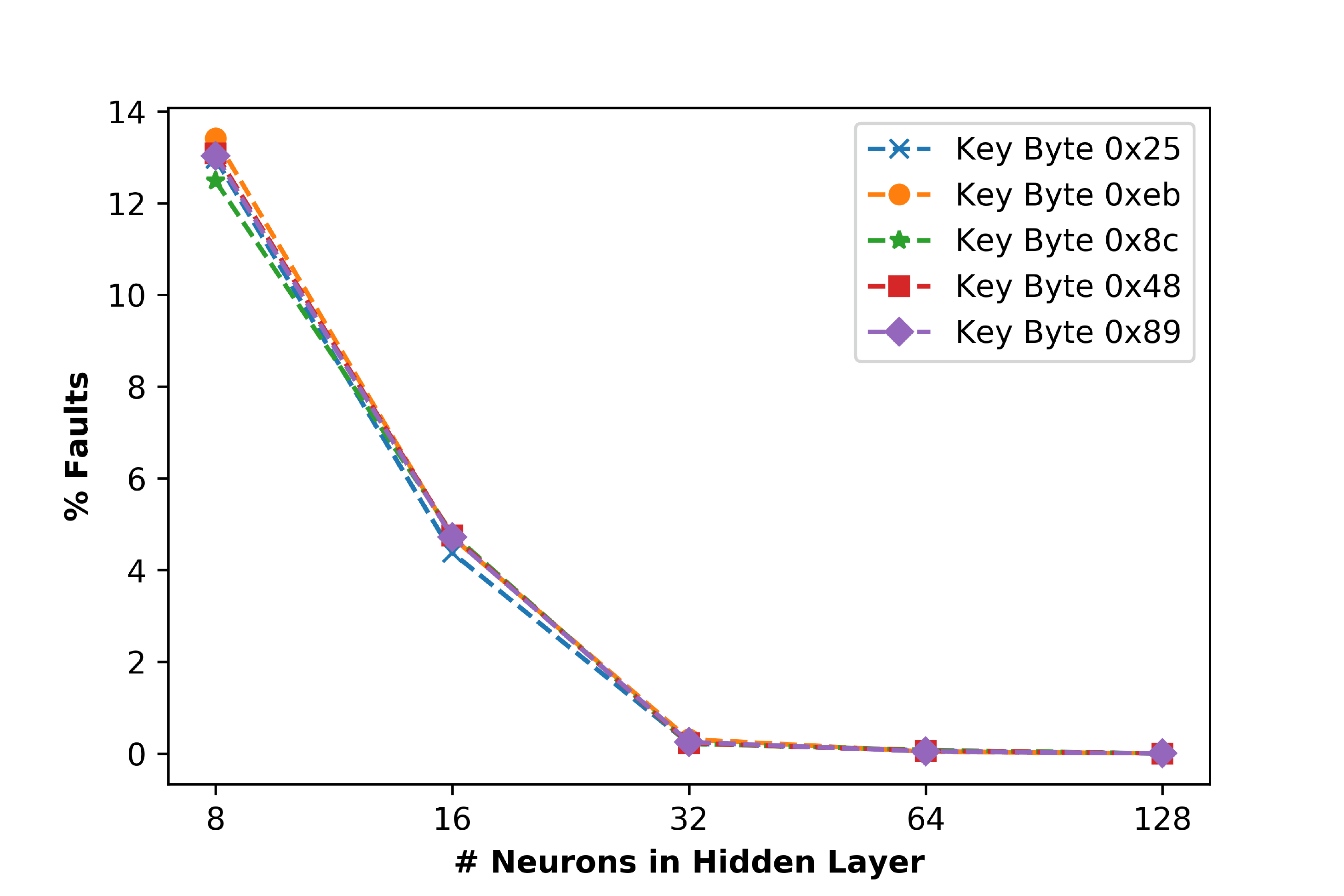}
	}
	\qquad
	\subfloat[Hidden Layer Bias]{
		\includegraphics[width=0.5\linewidth]{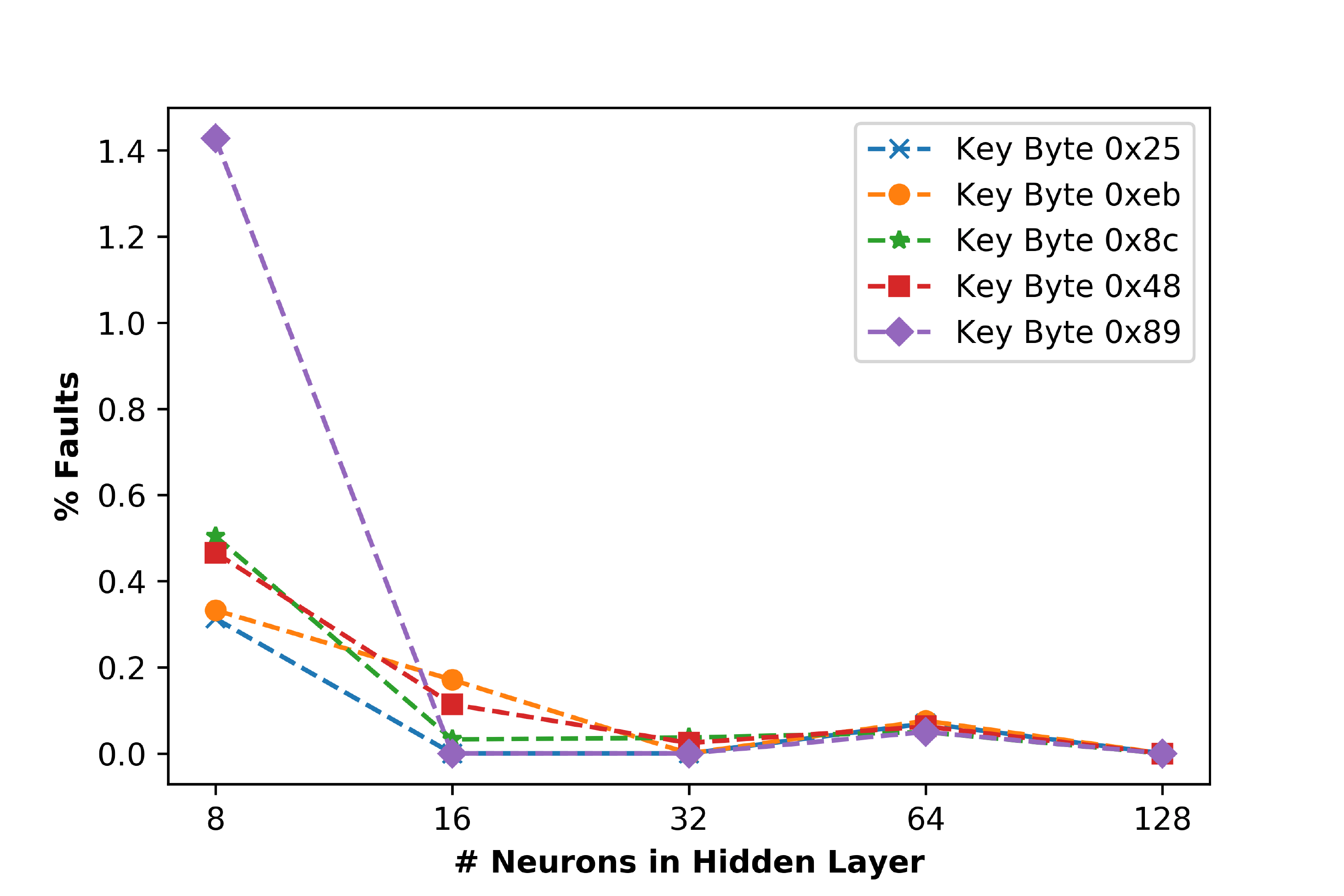}
	}
	\caption{Effect of Number of Neurons in Hidden Layer on Fault Tolerance\label{fig:fault_hidden}}
\end{figure}

Fig.~\ref{fig:fault_hidden} presents \% Faults for the previously mentioned five different NNs when faults are injected at the same parameter, as discussed before, considering different number of hidden layer neurons. We can see from the figure that with an increase in number of neurons the \% Faults for all the parameters gets decreased, thereby supporting the need of large number of neurons in the hidden layer. We have selected 128 neurons in the hidden layer for all our further experiments as it exhibits significantly high fault tolerance.

\begin{figure}[!t]
	\centering
	\subfloat[First Layer Weight]{
		\includegraphics[width=0.5\linewidth]{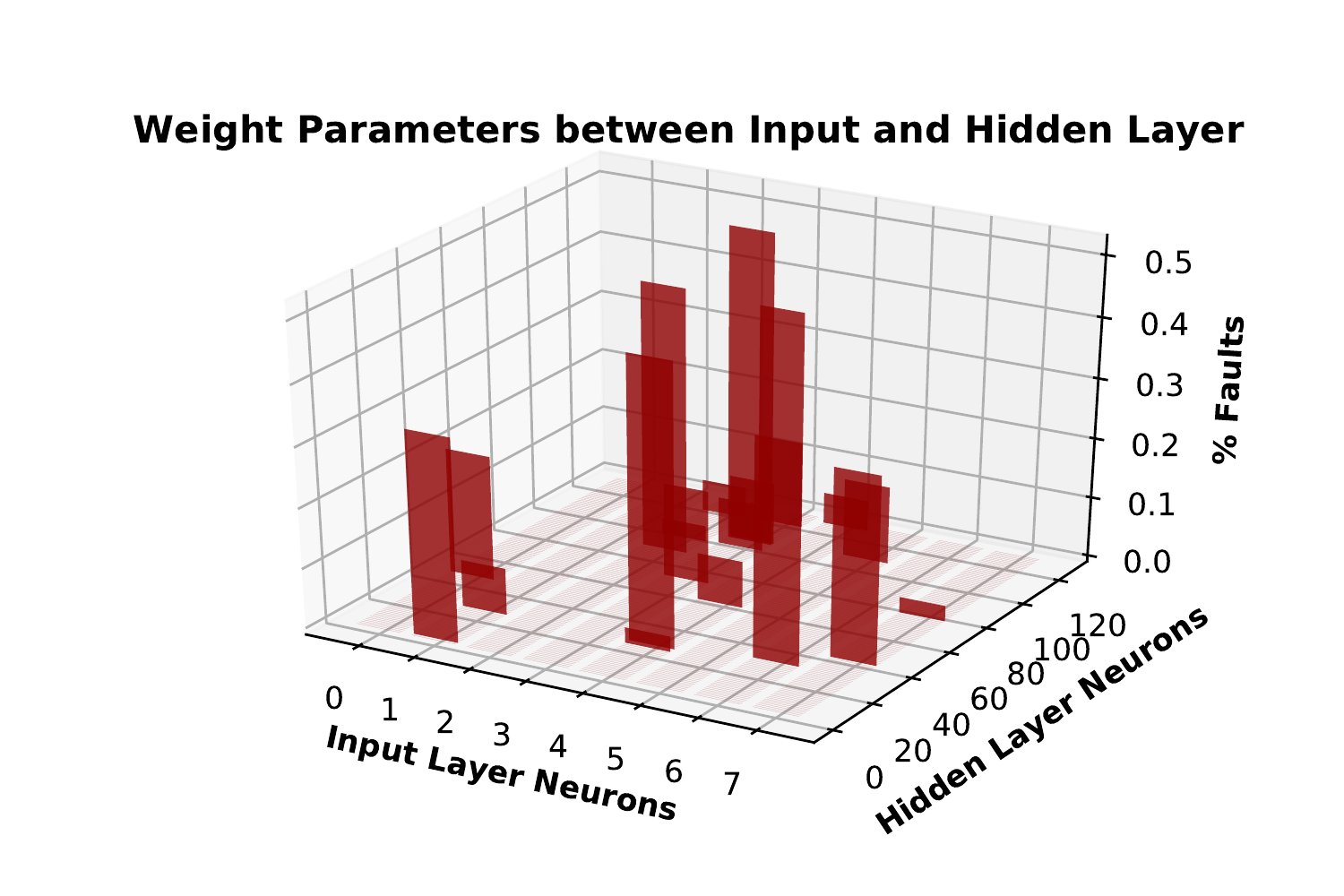}
	}
	\subfloat[Second Layer Weight]{
		\includegraphics[width=0.5\linewidth]{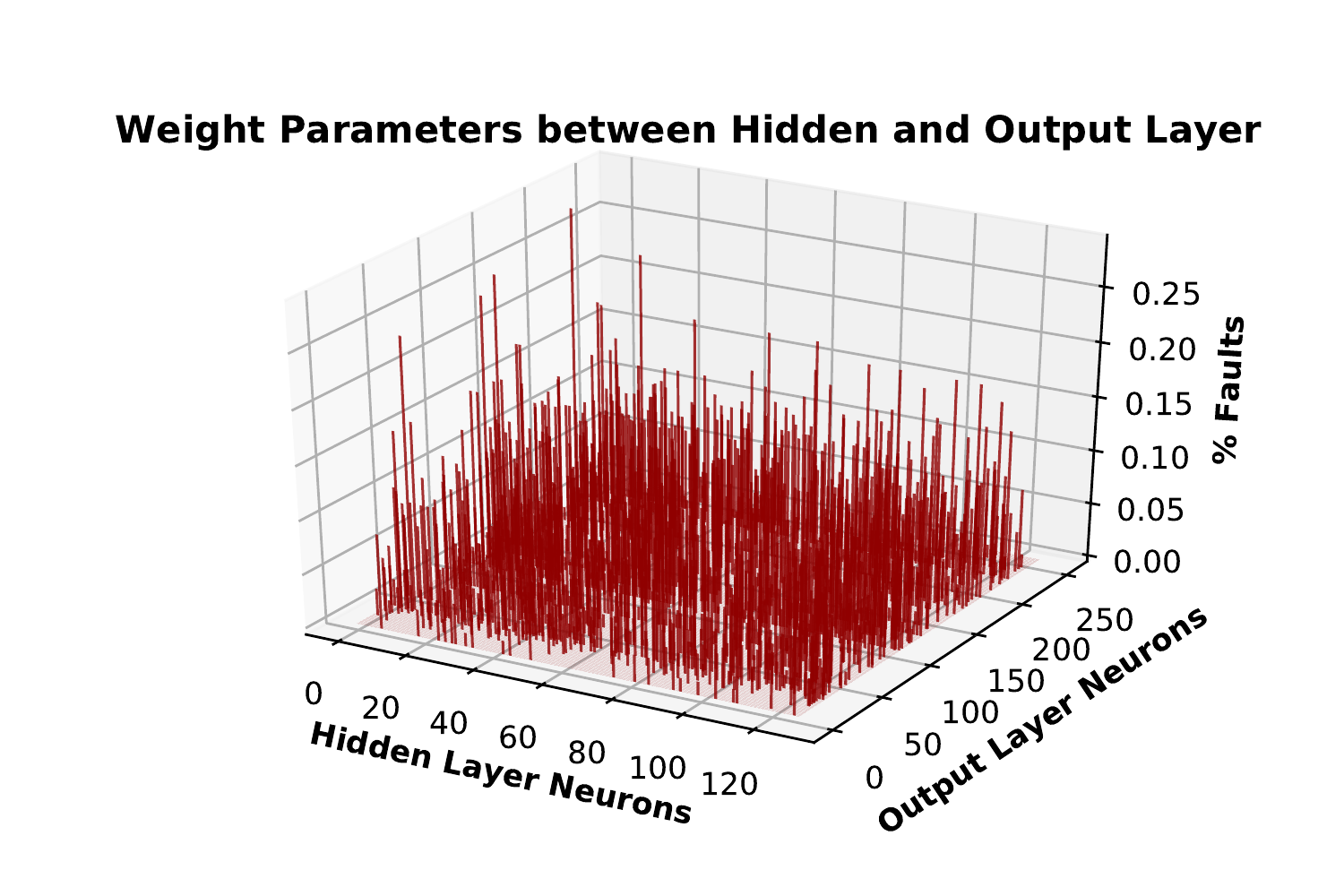}
	}
	\caption{Effect of each weight parameters on overall fault for the Standard Neural Network Architecture\label{fig:fault_parameter_old}}
\end{figure}

Fig.~\ref{fig:fault_parameter_old} presents the \% Fault Value for each parameters in First Layer and Second Layer Weights of the NN trained with secret key byte \texttt{0x25}. We can observe from the figure that most of the parameters in First Layer Weight are entirely fault resistant, whereas, faults in some of the weights can produce faulty outputs. However, for almost all of the parameters in Second Layer Weight, we can observe faulty outputs with properly chosen faults. 

Total number of faulty output for all possible combination of faulty parameters and inputs for NN with 128 hidden layer neurons is 6978 (147 due to First Layer Weight, 6810 due to Second Layer Weight, and 21 due to Hidden Layer Bias Parameters) out of 264,011,776 possibilities. Hence, PFT of this NN is $2.64\times10^{-3}\%$ faults. However, a cryptographic algorithm such as $f$ operation, as discussed previously, requires more fault tolerance from implementation. In order to make the computation more secure, we present an analytical way in the following subsection to enforce some constraints in NN parameters to increase PFT of the model.

\subsection{Conditions for Implementing Fully Fault-Tolerant Architecture}
\begin{figure}[!t]
		\centering
		\subfloat[]{\label{fig:fault_bias_2}
			\includegraphics[width=0.5\textwidth]{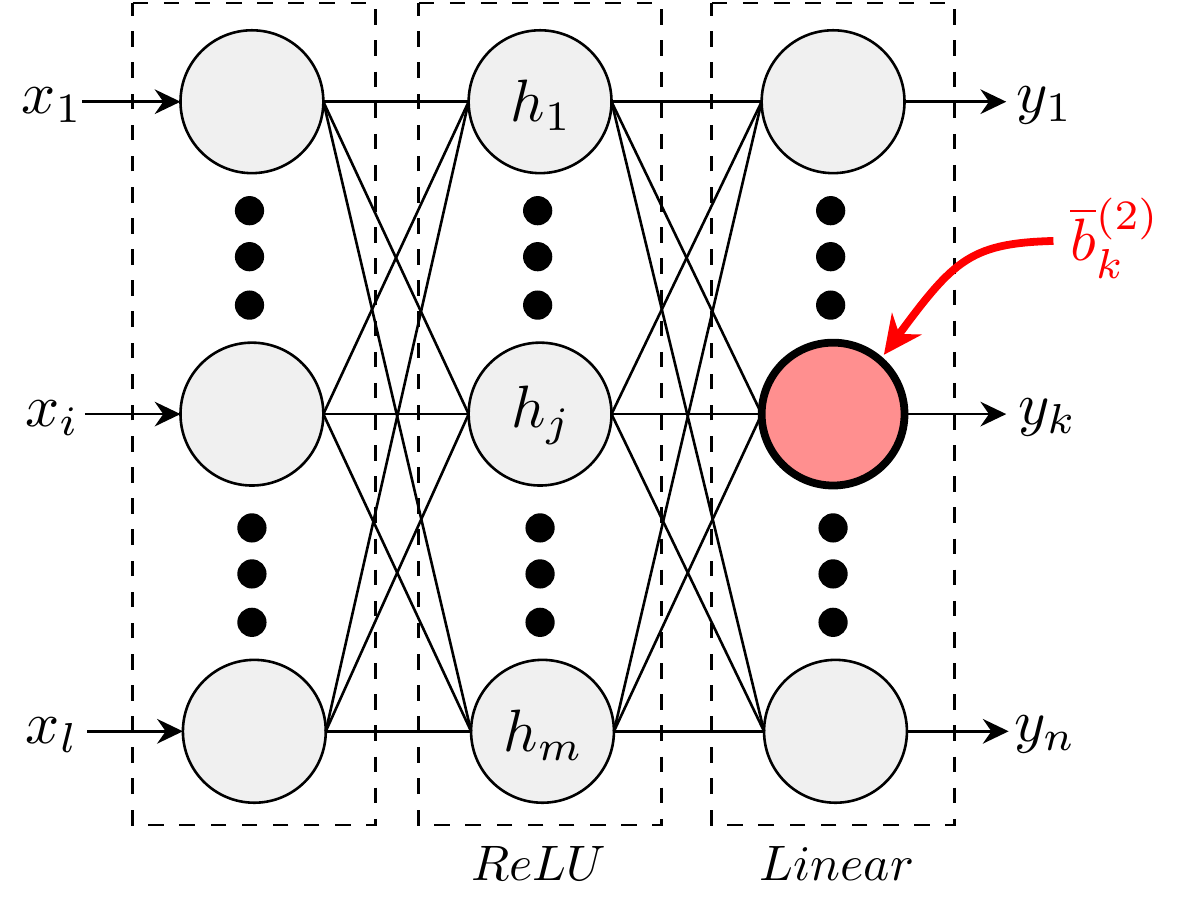}
		}
		\subfloat[]{\label{fig:fault_weight_2}
			\includegraphics[width=0.5\textwidth]{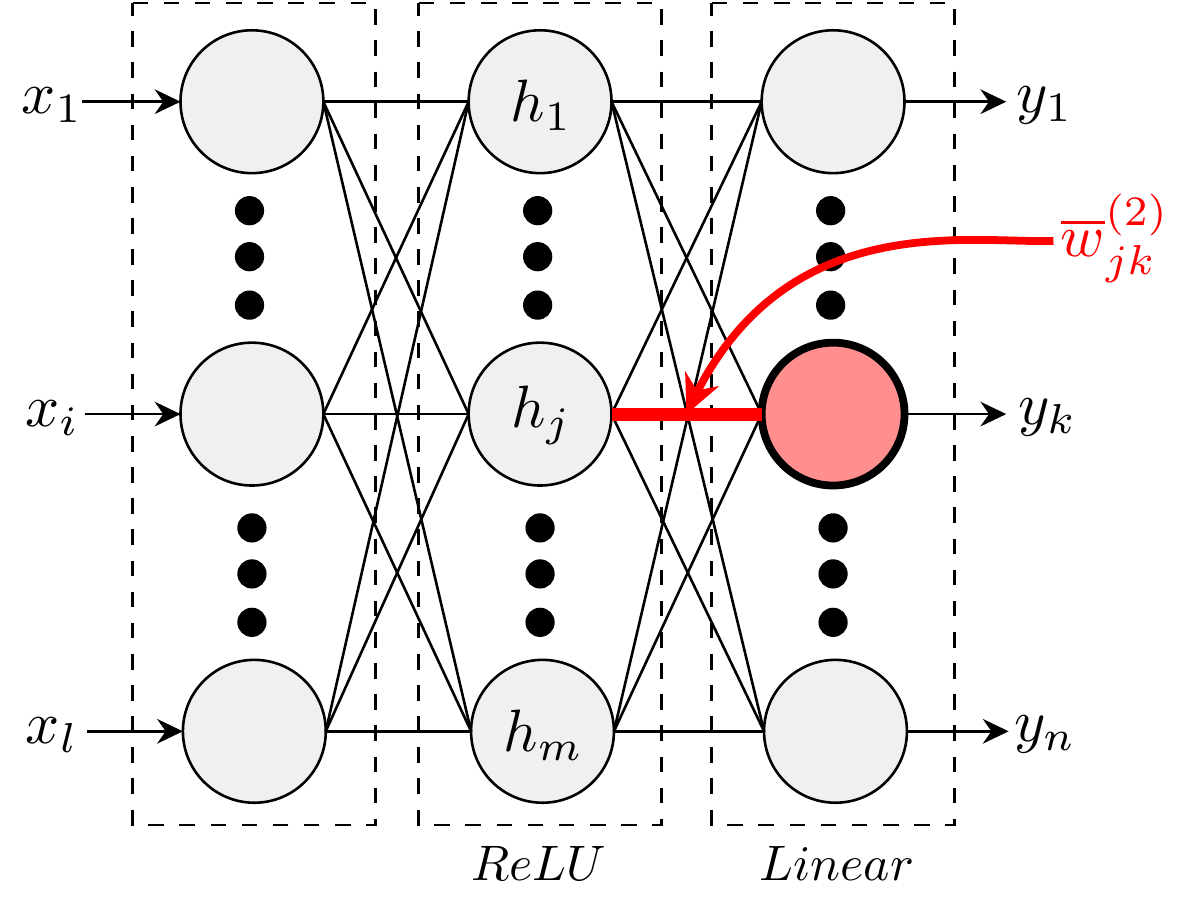}
		}
		\qquad
		\subfloat[]{\label{fig:fault_bias_1}
			\includegraphics[width=0.5\textwidth]{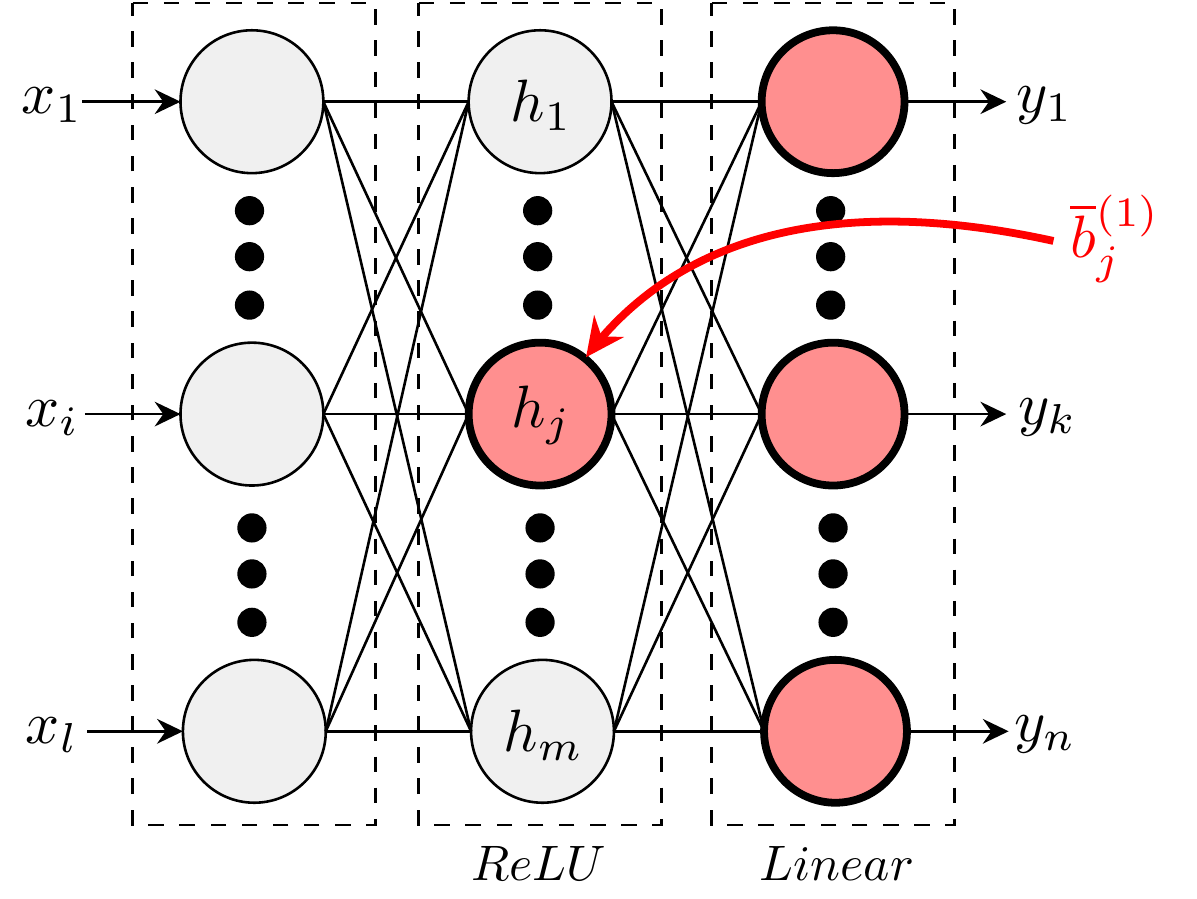}
		}
		\subfloat[]{\label{fig:fault_weight_1}
			\includegraphics[width=0.5\textwidth]{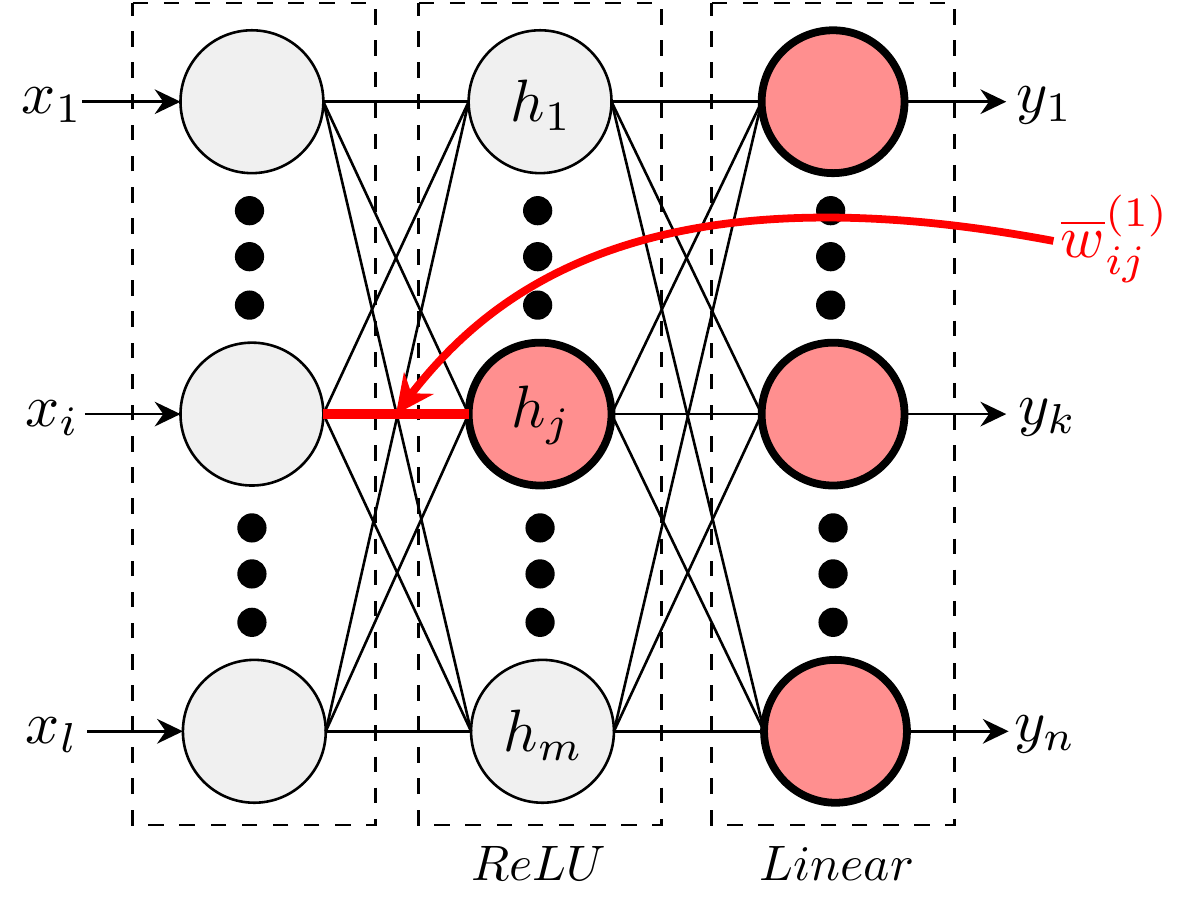}
		}
		\caption{\label{fig:fault_cases} Effect of Fault at Different Locations: a) Output Layer Bias, b) Weight Connecting Hidden and Output Layer, c) Hidden Layer Bias, and d) Weight Connecting Input and Hidden Layer. The \textcolor{red}{Red} coloured neurons signify the affected neurons because of the fault induction}
\end{figure}

An adversary can induce faults at any of the learned parameters of the trained NN model as discussed in Section~\ref{sec:fault_model}. In this subsection, we consider each of the parameters individually, as shown in Fig.~\ref{fig:fault_cases}, and evaluate the PFT of the NN architecture with respect to the induced fault.

\vspace{0.25cm}
\noindent \textbf{Case 1:} \underline{Fault injection in Output Layer Bias}

Let an adversary injects a fault and modifies $b_{f_2}^{(2)}$ by an amount $\delta$. Hence, the faulty value $\bar{b}_{f_2}^{(2)} = b_{f_2}^{(2)} \pm$ $\delta$. As a result of this fault, only the value of neuron $f_2$ at the output layer will be affected and all other neurons will be unaltered, which is shown in Fig.~\ref{fig:fault_bias_2}. Let the modified value of $y_{f_2}$ is $\bar{y}_{f_2}$. Hence, using Equation~\eqref{eq:linear},

\begin{equation*}
    \bar{y}_{f_2} = \sum_{j=1}^{m}h_jw_{j{f_2}}^{(2)} + \bar{b}_{f_2}^{(2)} = \sum_{j=1}^{m}h_jw_{j{f_2}}^{(2)} + b_{f_2}^{(2)} \pm \delta = y_{f_2} \pm \delta
\end{equation*}

Let for a particular input $x^c = (x_1^c, x_2^c, \dots, x_l^c)$, the correct class belongs to the output node $c$, i.e, the value of $y_c$ is maximum because of the $\texttt{ArgMax}$ function. Since the adversary does not induce any fault in any $w_{ij}^{(1)}$ and $b_j^{(1)}$, all the values for the neurons in the hidden layer will remain unchanged. We consider following two scenarios for analyzing the fault tolerance of the network.

\begin{enumerate}[label=(\alph*)]
    \item \textbf{$c \neq f_2$; i.e., fault affects any node apart from the correct node}: Misclassification will happen due to the fault when $\bar{y}_{f_2} > y_c$, as it will then classify input $x^c$ to class $f_2$ instead of the correct class $c$ due to the $\texttt{ArgMax}$ function. Hence, the effect of fault has to increase the value of $y_{f_2}$, i.e., $y_{f_2} + \delta > y_c$. All other effect of faults which decreases the value of $y_{f_2}$ will have no impact on NN decision. Hence, the NN will be fault tolerant if $y_{f_2} + \delta < y_c \implies \delta < y_c - y_{f_2}$. 

    \item \textbf{$c = f_2$; i.e., fault affects the correct node}: Misclassification, in this case, will happen when $\bar{y}_{f_2} < y_r$, for some neuron $r$ in the output layer. Hence, the effect of fault has to decrease the value of $y_{f_2}$, i.e., $y_{f_2} - \delta < y_r$ for some $r$. All the effect of faults which increases the value of $y_{f_2}$ will have no impact on the NN decision. Hence, the NN will be fault tolerant if $y_{f_2} - \delta > y_r \implies \delta < y_{f_2} - y_r$.
\end{enumerate}

Therefore, the condition for fault-tolerant NN when the fault is injected in Output Layer Bias is

\begin{equation}\label{eq:out_bias}
\delta <
\begin{cases}
      y_c - y_{f_2}, & \text{if}\ c \neq f_{2} \\
      y_{f_2} - y_r, & \text{otherwise, for all $r$}
    \end{cases}
\end{equation}

\vspace{0.25cm}
\noindent \textbf{Case 2:} \underline{Fault injection in Weight between Hidden-Output Layer}

Let an adversary injects a fault and modifies $w_{f_1f_2}^{(2)}$ by an amount $\delta$. Hence, the faulty value $\bar{w}_{f_1f_2}^{(2)} = w_{f_1f_2}^{(2)} \pm$ $\delta$. The situation is shown in Fig.~\ref{fig:fault_weight_2}. Proceeding with a similar approach like \textbf{Case 1}, the condition for fault-tolerant NN in this case is

\begin{equation}\label{eq:out_weight}
\delta <
\begin{cases}
      \frac{y_c - y_{f_2}}{h_{f_1}}, & \text{if}\ c \neq f_{2}\\
      \frac{y_{f_2} - y_r}{h_{f_1}}, & \text{otherwise, for all $r$}
    \end{cases}
\end{equation}

\vspace{0.25cm}
\noindent \textbf{Case 3:} \underline{Fault injection in Hidden Layer Bias}

Let an adversary injects a fault and modifies $b_{f_1}^{(1)}$ by an amount $\delta$. Hence, the faulty value $\bar{b}_{f_1}^{(1)} = b_{f_1}^{(1)} \pm \delta$. As a result of this fault, value of neuron $f_1$ at the hidden layer will be affected and all the neurons in the output layer will be affected as the value $h_{f_1}$ is propagated to all  neurons in the output layer, which is shown in Fig.~\ref{fig:fault_bias_1}.

Let the modified value of $h_{f_1}$ after the fault injection is $\bar{h}_{f_1}$. Hence,

\begin{equation*}
    \bar{h}_{f_1} = \sum_{i=1}^{l}x_iw_{i{f_1}}^{(1)} + \bar{b}_{f_1}^{(1)} = \sum_{i=1}^{l}x_iw_{i{f_1}}^{(1)} + b_{f_1}^{(1)} \pm \delta = h_{f_1} \pm \delta
\end{equation*}

Let for a particular input $x^c = (x_1^c, x_2^c, \dots, x_l^c)$, the correct class belongs to the output node $c$, i.e, the value of $y_c$ is maximum because of the $\texttt{ArgMax}$ function. The modified value of $y_c$ is given by,

\begin{align*}
    \bar{y}_{c} &= \sum_{\substack{j=1 \\ j \neq f_1}}^{m}h_jw_{jc}^{(2)} + \bar{h}_{f_1}w_{f_1c}^{(2)} + b_{c}^{(2)} = \sum_{\substack{j=1 \\ j \neq f_1}}^{m}h_jw_{jc}^{(2)} + (h_{f_1} \pm \delta)w_{f_1c}^{(2)} + b_{c}^{(2)}\\
    &= \sum_{\substack{j=1 \\ j \neq f_1}}^{m}h_jw_{jc}^{(2)} + h_{f_1}w_{f_1c}^{(2)} + b_{c}^{(2)} \pm \delta w_{f_1c}^{(2)} = y_c \pm \delta w_{f_1c}^{(2)}
\end{align*}

The modified value for any other neuron $r$ can be derived as the same way and is given by: $\bar{y}_r = y_r \pm \delta w_{f_1r}^{(2)}$. The NN will be fault tolerant if $\bar{y}_c > \bar{y}_r$ for all $r$, i.e., $y_c \pm \delta w_{f_1c}^{(2)} > y_r \pm \delta w_{f_1r}^{(2)}$. Hence the condition for fault tolerance in this case is

\begin{equation}\label{eq:hidden_bias}
    \delta < \pm \frac{y_c - y_r}{w_{f_1r}^{(2)} - w_{f_1c}^{(2)}}
\end{equation}

\vspace{0.25cm}
\noindent \textbf{Case 4:} \underline{Fault injection in Weight between Input-Hidden Layer}

Let an adversary injects a fault and modifies $w_{f_0f_1}^{(1)}$ by an amount $\delta$. Hence, the faulty value $\bar{w}_{f_0f_1}^{(1)} = w_{f_0f_1}^{(1)} \pm \delta$. The situation is shown in Fig.~\ref{fig:fault_weight_1}. Proceeding with a similar approach like \textbf{Case 3}, the condition for fault tolerance NN in this case is

\begin{equation}\label{eq:hidden_weight}
    \delta < \pm \frac{y_c - y_r}{w_{f_0f_1}^{(1)}w_{f_1r}^{(2)} - w_{f_0f_1}^{(1)}w_{f_1c}^{(2)}}
\end{equation}

In the following subsection, we present an analysis on learning AES SBox using the NN as mentioned before with the four constraints presented in this section.

\subsection{Fault Tolerance of AES SBox learned with Modified Neural Network}~\label{sec:highest_pft}
We have used the same NN architecture as mentioned in Section~\ref{sec:dataset_topology}, i.e., with 8 input layer neurons, 128 hidden layer neurons, and 256 output layer neurons considering integer weights, but with an additional constraints on weight and bias parameters. The constraint on parameters have been imposed by \texttt{L2}-\texttt{Regularization}, which adds penalty to the cost function of the NN while training, restricting the parameters to be constrained within a fixed boundary. We have tried to replicate different models for which the conditions mentioned in Equation~\eqref{eq:out_bias}, Equation~\eqref{eq:out_weight}, Equation~\eqref{eq:hidden_bias}, and Equation~\eqref{eq:hidden_weight} holds. We found out a model with significantly higher PFT than the standard implementation having no constraints on the parameters.

\begin{figure}[!t]
	\centering
	\subfloat[First Layer Weight]{
		\includegraphics[width=0.5\linewidth]{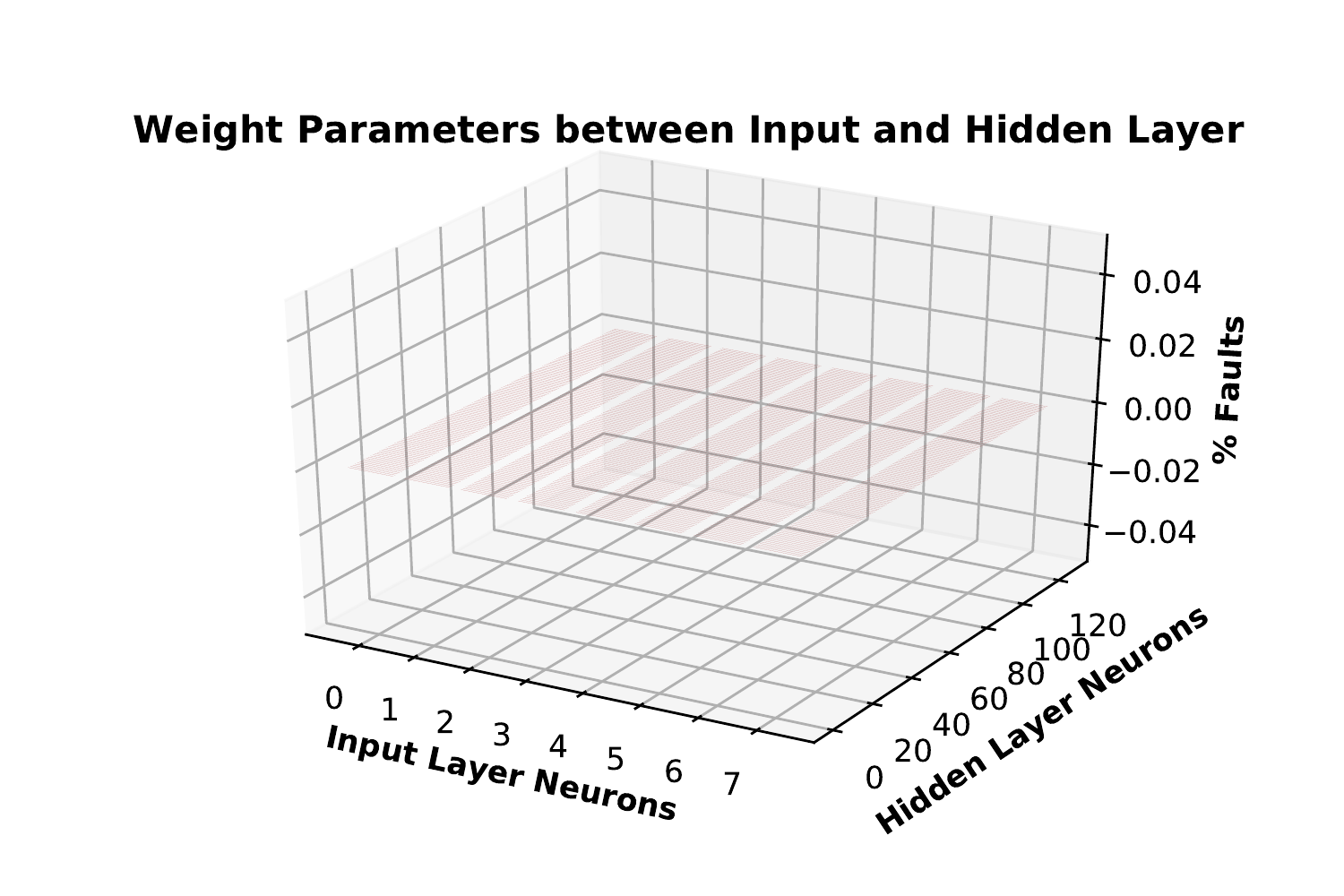}
	}
	\subfloat[Second Layer Weight]{
		\includegraphics[width=0.5\linewidth]{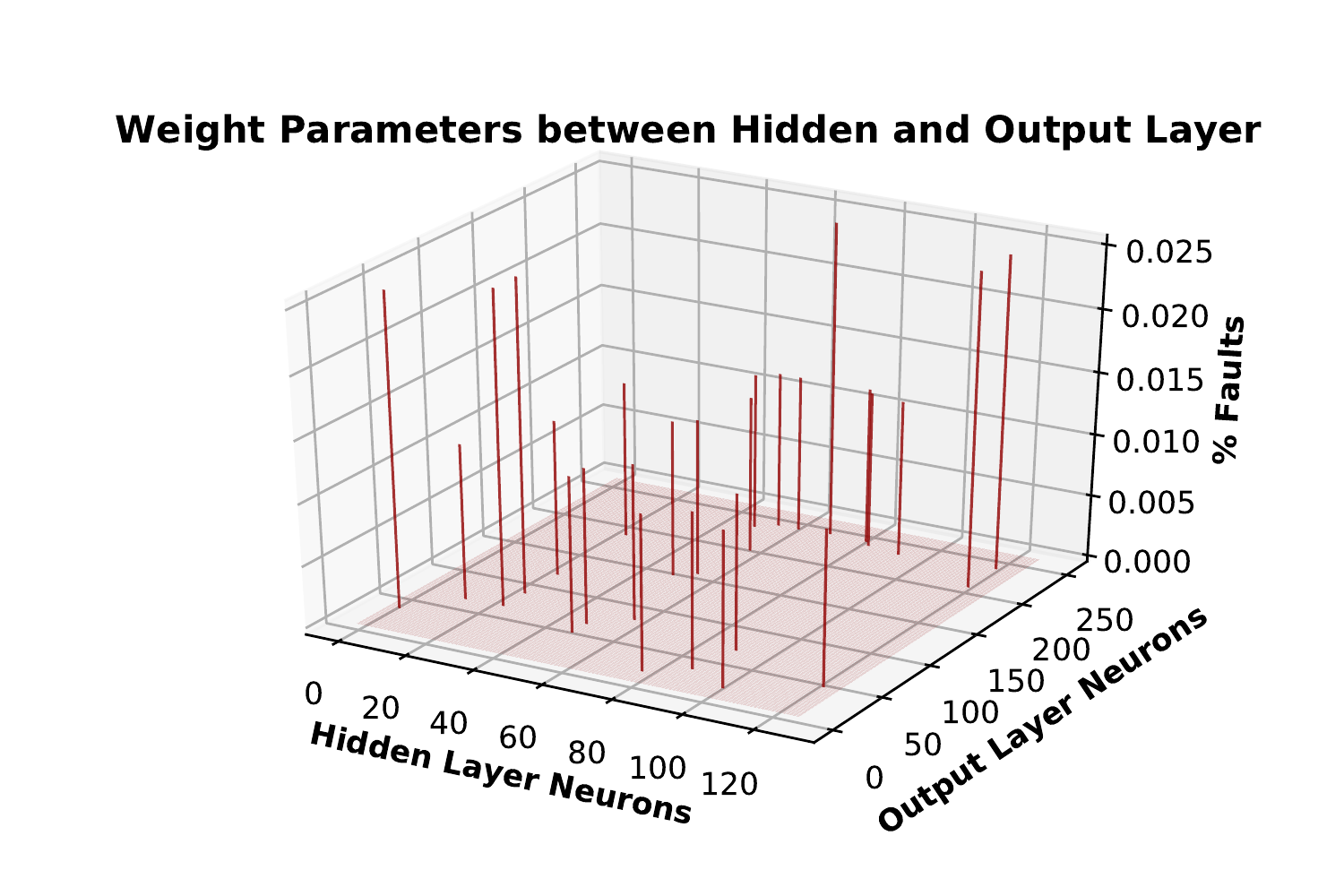}
	}
	\caption{Effect of each weight parameters on overall fault for the Modified Neural Network Architecture\label{fig:fault_parameter_new}}
\end{figure}

Fig.~\ref{fig:fault_parameter_new} presents the \% Fault Value for each parameters in First Layer Weights and Second Layer Weights for such a NN trained with the secret key byte \texttt{0x25}. We can easily observe from the figure that the parameters in First Layer Weights are completely fault tolerant while there are very few faulty outputs due to fault injection in the Second Layer Weights. In the modified implementation, all the bias parameters are completely fault tolerant, which effectively prevents the \emph{Single Bias Attack} mentioned in~\cite{liu2017fault}. In addition to it, number of sensitive weight parameters gets significantly reduced in the modified implementation. As a result the \emph{Gradient Descent Attack}~\cite{liu2017fault} becomes more difficult to mount as the searching complexity of sensitive weight parameters gets increased significantly. In addition to it, a designer already has information about the sensitive weights, which he can protect instead of all the weights exhaustively making the attack more difficult.

The total number of faulty output for all possible combination of faulty parameters and inputs, in this case, is 32 (due to Second Layer Weight Parameters only) out of 264,011,776 possibilities. Hence, the PFT of this NN is $1.21\times10^{-5}\%$ faults. We can easily observe that the PFT of the modified model is \textbf{218$\times$} better than the previous model which had no constraints on the parameters.

\section{Implementation on an FPGA}
In this section, we focus on the overhead of different NN architectures for execution of $f$ operation. To the best of our knowledge, the study of implementation overhead for achieving a higher degree of fault tolerance is not present in literature in the context of NNs.

From Equation~\eqref{eq:relu} and Equation~\eqref{eq:linear}, we can observe that the computation of a neuron can be expressed with simple multiplication and addition operations. This can be efficiently executed by Digital Signal Processor (DSP) blocks of modern FPGAs which supports fast multiplication and addition of integers. This motivates our choice of shifting to integer valued weights from floating point. The architecture of a single NN-based $f$ operation can be realised either in an iterative or in a parallel fashion depending on area, time constraints. In our proposed implementation, we have opted for an iterative architecture to have a compact implementation, which is validated on an Artix-7 FPGA (7a35t-cpg236).

\begin{figure}[!t]
\centering
\includegraphics[width=0.75\textwidth]{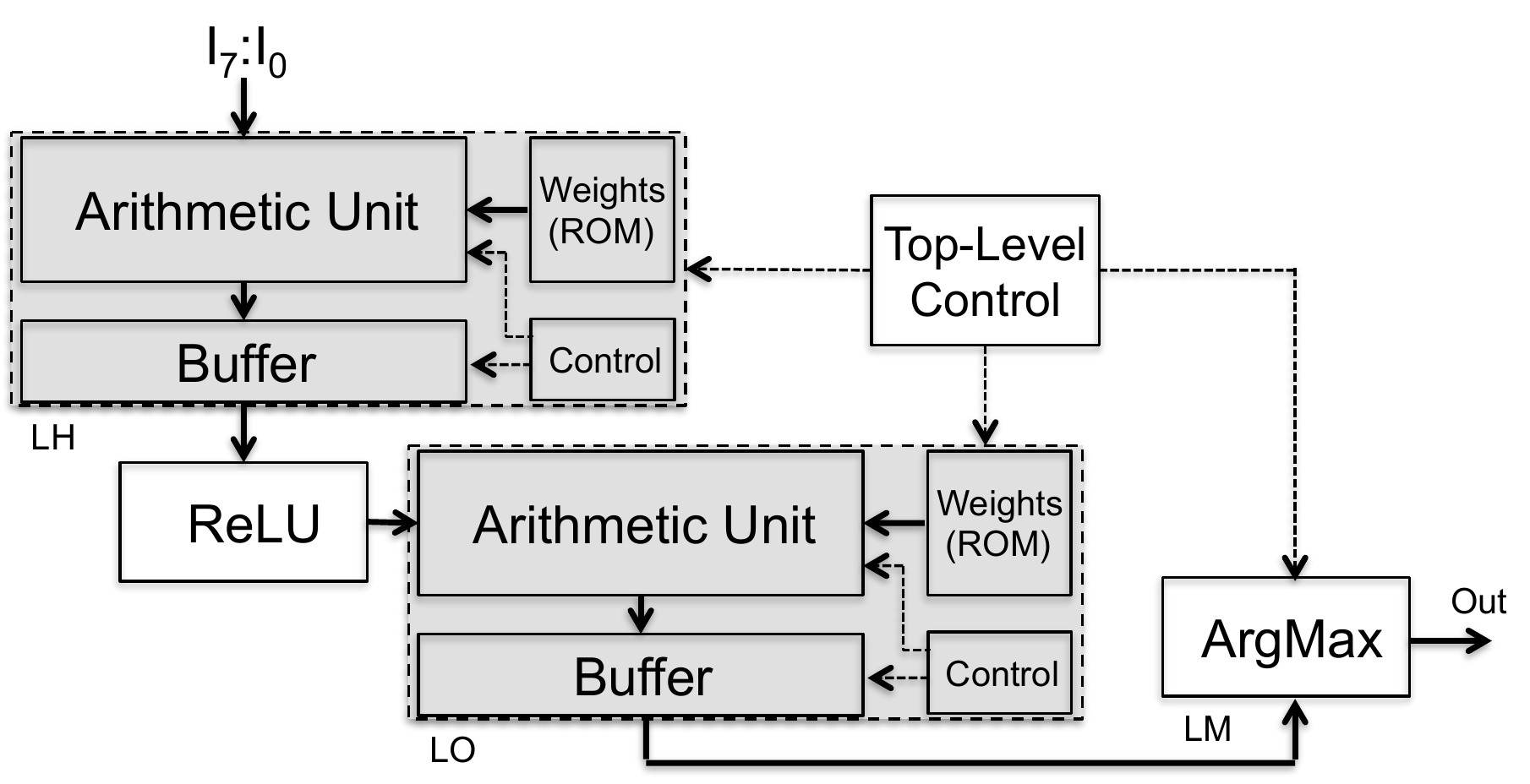}
\caption{Top Level Architecture of Hidden and Output Layer}
\label{fig:TOP_DSGN}
\end{figure}

A top-level description of the complete architecture corresponding to the 8-128-256 NN is provided in Fig.~\ref{fig:TOP_DSGN}. The hidden layer computation is performed by the LH and ReLU block. Output of ReLU block is forwarded to the LO block for the output layer computation. The LO block operates in the same way as LH block, but the data widths are different. 
Output of the LO block is forwarded to \texttt{ArgMax} block which finds the index of maximum value and finally produces an output of the design. Top-level control block controls the operations of these blocks, which is designed using a Finite State Machine.

\begin{figure}[!t]
\centering
\includegraphics[width=0.8\textwidth]{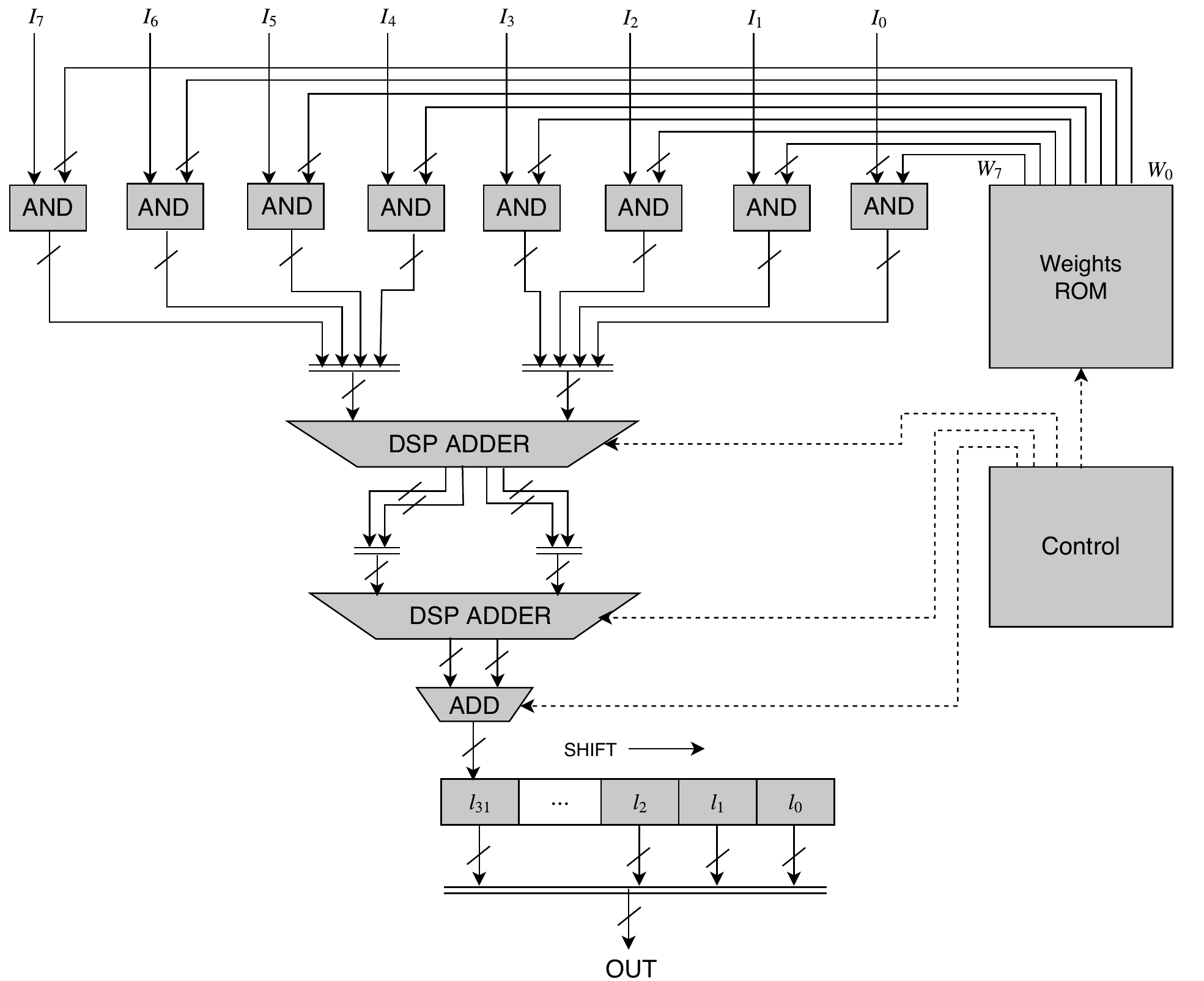}
\caption{Top Level Architecture of the Hidden Layer}
\label{fig:TOP_FSTLYR}
\end{figure}

Fig.~\ref{fig:TOP_FSTLYR} provides a broad overview of the architecture of LH block. Inputs to this layer (hidden layer) are the eight-bit values of the inputs to $f$ operation. Since the input bits are either 0 or 1, no multiplication of the weight values with the input is required. The weight values corresponding to bit 1 are added, and the bias is added finally to get the output. The selection of weight values is made using multiple \texttt{AND} gates as depicted in Fig.~\ref{fig:TOP_FSTLYR}.  Block RAMs are used to store the weights and bias values. Addition of the weight values is done using two DSP adders with appropriate zero padding. Final addition uses normal LUT-based adder as the width of the inputs are small ($9$ bits). The iterative architecture requires a total of 128 iterations to compute the output of all the 128 neurons. The new value is inserted into the buffer and is shifted right after each iteration. The output is directly taken from the buffer register when 128 iterations are complete.

\begin{figure}[!t]
\centering
\includegraphics[width=\textwidth]{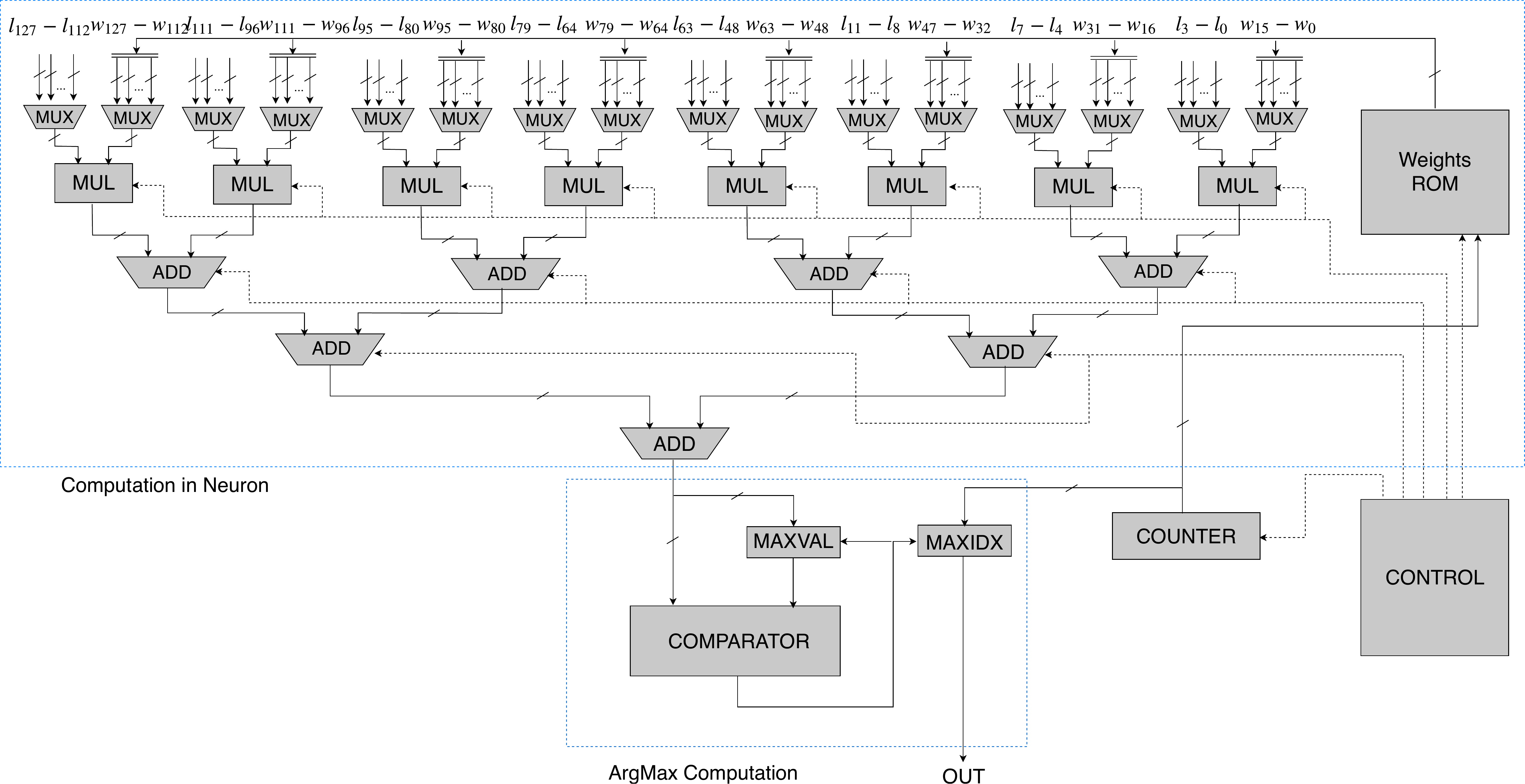}
\caption{Architecture of the Output Layer with \emph{\texttt{ArgMax}} Function}
\label{fig:ARCH_SCNDLYR}
\end{figure}

Fig.~\ref{fig:ARCH_SCNDLYR} shows the architectural diagram of the output layer along with the \texttt{ArgMax} block, which has a similar structure like LH block, and also operates in iterative manner. However, this layer has a large number of inputs compared to the hidden layer and a limited number of multipliers (to keep resource usage low). Hence, multiplexers have been utilised to select the inputs for each execution of the computation unit. Each of the multipliers has been constructed using multiple DSP blocks and is capable of multiplying 20-bit values. DSP and LUT-based adders are also used in the multiplier. Eight parallel instantiations have been used for the computation, and hence, computation of one neuron takes $128/8=16$ executions of the arithmetic unit. Total 256 iterations are required for the 256 neurons. The output of each iteration is forwarded to the \texttt{ArgMax} block which finds the index of the maximum value. Finally, the index of the maximum value is stored in a register of \texttt{ArgMax} block (LM) which is passed to the output of the design after LO block computation is finished.

Table~\ref{tab:RESCOMP} provides a comparative summary of different NN architectures performing $f$ operation on Artix-7 FPGA. The complete design (8-128-256), which exhibits the highest PFT, operates at a maximum operating frequency of 128.66 MHz (critical path of 7.77 ns). We also report the resource usage for a standard AES SBox in the last row of Table~\ref{tab:RESCOMP}. We conclude from the table that, as the number of neurons in the hidden layer increases, resource and timing requirement of the designs also increases. This is the overhead penalty that a designer needs to pay to achieve significantly high PFT as shown in Section~\ref{sec:highest_pft}. The resource usage for LUT-based AES SBox is extremely low, and the maximum operating frequency is also higher. However, this implementation has zero fault tolerance as the entire architecture is deterministic.

\begin{figure}[!t]
\centering
\includegraphics[width=0.7\linewidth]{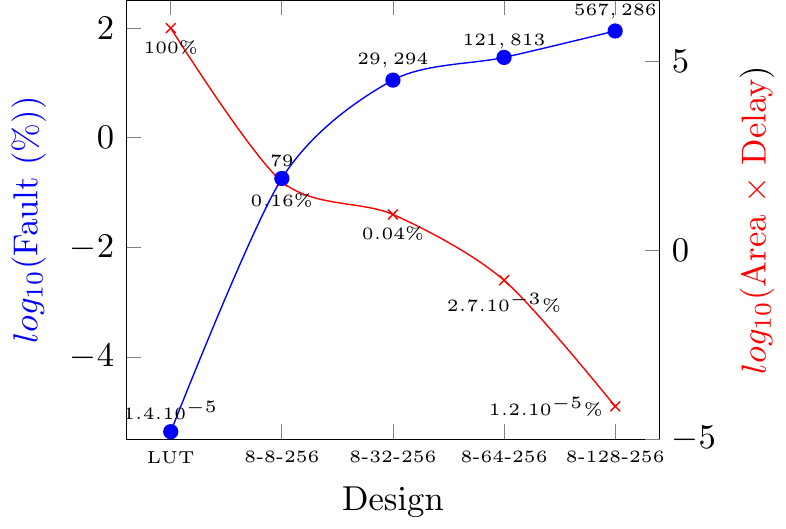}
\caption{Fault Tolerance vs. Resource Overhead trade-off}
\label{fig:summary}
\end{figure}

We have performed on-chip validation of the proposed modified learning strategy by injecting faults into the 8-128-256 architecture. We used the clock glitch method to inject faults at a single location into the design at different time instances during its operation. The final result of fault occurrences is consistent with our simulation-based experiments. Fig.~\ref{fig:summary} provides an overview of the on-chip experimental validations of our idea focusing on the trade-off between PFT and implementation overhead. It is clear from the figure that the LUT-based AES SBox, though is least in terms of area-delay product, produces highest number of faulty outputs than any other implementation. The 8-8-256 architecture without any constraints provides a better PFT but with higher resource utilization. However, when we modify the learning strategy with the constraints and increase number of neurons in the hidden layer, we can observe that the PFT increases significantly. We can see that the PFT is maximum for 8-128-256 architecture, though it comes with a higher overhead. The results validate our idea that the proposed learning strategy can enhance the fault tolerance of NNs.

\begin{table}[!t]
\centering
\caption{Post Place\&Route Resource Utilisation for Artrix-7 FPGA for Different Implementations}
\label{tab:RESCOMP}
\resizebox{\columnwidth}{!}{%
\begin{tabular}{|c|c|c|c|c|c|c|c|c|c|}
\hline
\textbf{Design} & \textbf{\begin{tabular}[c]{@{}c@{}}\#Slice\\ (\%)\end{tabular}} & \textbf{\begin{tabular}[c]{@{}c@{}}\#LUT\\ (\%)\end{tabular}} & \textbf{\begin{tabular}[c]{@{}c@{}}\#Register\\ (\%)\end{tabular}} & \textbf{\begin{tabular}[c]{@{}c@{}}\#DSP\\ (\%)\end{tabular}} & \textbf{\begin{tabular}[c]{@{}c@{}}\#BRAM\\ (\%)\end{tabular}} & \textbf{\begin{tabular}[c]{@{}c@{}}Freq.\\ (MHz)\end{tabular} } & \textbf{\begin{tabular}[c]{@{}c@{}}\#Clock\\ Cycle\end{tabular}} & \textbf{\begin{tabular}[c]{@{}c@{}}Delay\\ (us)\end{tabular}} & \textbf{\begin{tabular}[c]{@{}c@{}}\\ \% Faults \end{tabular}}\\ \hline\hline
\textbf{8-8-256}\footnote{Without any constraints on the parameters.} & \begin{tabular}[c]{@{}c@{}}127\\ (1.55)\end{tabular} & \begin{tabular}[c]{@{}c@{}}324\\ (1.56)\end{tabular} & \begin{tabular}[c]{@{}c@{}}199\\ (0.47)\end{tabular} & \begin{tabular}[c]{@{}c@{}}33\\ (36.67)\end{tabular} & \begin{tabular}[c]{@{}c@{}}5\\ (10)\end{tabular} & 151.95 & 1350 & 8.88 & 0.16 \\ \hline
\textbf{8-32-256} & \begin{tabular}[c]{@{}c@{}}341\\ (4.18)\end{tabular} & \begin{tabular}[c]{@{}c@{}}934\\ (4.49)\end{tabular} & \begin{tabular}[c]{@{}c@{}}951\\ (2.29)\end{tabular} & \begin{tabular}[c]{@{}c@{}}33\\ (36.67)\end{tabular} & \begin{tabular}[c]{@{}c@{}}4\\ (8)\end{tabular} & 149.43 & 25576 & 171.15 & 0.04 \\ \hline
\textbf{8-64-256}\footnote{\% Faults value for this network is equivalent to the 8-128-256 architecture without any constraints as mentioned in Section~\ref{sec:design_standard}, which signifies that a smaller network with the proposed constraints on the parameters can achieve PFT of large network without any constraint.} & \begin{tabular}[c]{@{}c@{}}427\\ (5.24)\end{tabular} & \begin{tabular}[c]{@{}c@{}}978\\ (4.70)\end{tabular} & \begin{tabular}[c]{@{}c@{}}1007\\ (2.43)\end{tabular} & \begin{tabular}[c]{@{}c@{}}33\\ (36.67)\end{tabular} & \begin{tabular}[c]{@{}c@{}}6\\ (12)\end{tabular} & 141.40 & 49352 & 349.01 & $2.7\times10^{-3}$ \\ \hline
\textbf{8-128-256} & \begin{tabular}[c]{@{}c@{}}601\\ (7.37)\end{tabular} & \begin{tabular}[c]{@{}c@{}}1442\\ (6.93)\end{tabular} & \begin{tabular}[c]{@{}c@{}}1657\\ (3.98)\end{tabular} & \begin{tabular}[c]{@{}c@{}}33\\ (36.67)\end{tabular} & \begin{tabular}[c]{@{}c@{}}9\\ (18)\end{tabular} & 128.66 & 96910 & 753.18 & $1.2\times10^{-5}$ \\ \hline\hline
\textbf{\begin{tabular}[c]{@{}c@{}}LUT-based\\\end{tabular}} & \begin{tabular}[c]{@{}c@{}}24\\ (0.29)\end{tabular} & \begin{tabular}[c]{@{}c@{}}80\\ (0.40)\end{tabular} & \begin{tabular}[c]{@{}c@{}}17\\ (0.40)\end{tabular} & \begin{tabular}[c]{@{}c@{}}0\\ (00)\end{tabular} & \begin{tabular}[c]{@{}c@{}}0\\ (00)\end{tabular} & 259.80 & 1 & \begin{tabular}[c]{@{}c@{}}$3.85$\\ $\times10^{-3}$\end{tabular} & 100 \\ \hline
\end{tabular}
}
\end{table}

\section{Conclusion}
In this paper, we propose design of cryptographic primitives using NN for high fault tolerance.
As a case study, we showed practical implementation of AES Sbox (with key addition).
We then propose a technique to further boost the fault tolerance of designed primitive with some analytically derived constraints on the network parameters, which increases the complexity of fault injection based attacks on NNs. A tailored implementation strategy for the NN using integer weights is validated on an FPGA. We show that fault tolerance can be scaled up albeit with higher area/delay overhead.

\bibliographystyle{splncs04}
\bibliography{bib}
\end{document}